\documentclass[11pt,a4paper]{article}
\usepackage[hyperref]{emnlp2018}
\usepackage{times}
\usepackage{latexsym}

\usepackage{url}
\usepackage{graphicx}     
\aclfinalcopy 

\title{Recent advances in conversational NLP :\\
Towards the standardization of Chatbot building 
}

\author{Maali Mnasri \\
  Opla / Clermont-Ferrand, France \\
  {\tt maali@opla.ai} \\}

\date{}

\begin{document}
\maketitle
\begin{abstract}
Dialogue systems have become recently essential in our life. Their use is getting more and more fluid and easy throughout the time. This boils down to the improvements made in NLP and AI fields. In this paper, we try to provide an overview to the current state of the art of dialogue systems, their categories and the different approaches to build them. We end up with a discussion that compares all the techniques and analyzes the strengths and weaknesses of each. Finally, we present an opinion piece suggesting to orientate the research towards the standardization of dialogue systems building.

\end{abstract}

\section{Introduction}

Conversational agents or dialogue systems, referred also as chatbots by the media and the industrials, have become very common in our everyday life. We encounter them, for example, in our mobile phones as personal assistants or in e-commerce websites as selling bots. These systems are intended to carry coherent conversations with humans in natural language text or speech or even both. Developing intelligent conversational agents is still an unresolved research problem that raises many challenges in the artificial intelligence community.
We try through this work to identify the different existing algorithms to build chatbots. We also classify the predominant approaches and compare them according to the final use case while stating each approach strengths and weaknesses. We aim at uncovering the issues related to this task which can help researchers choose the future directions in conversational NLP.\\
Although the first chatbots have been developed many years ago, this field has never been more focused on than these last years. This can be explained by the recent development on AI and NLP technologies as well as the data availability. The current scientific and technological landscape is starting to get crowded with the variety of methods to build chatbots \cite{gilbert2019learning, haponchyk2018supervised, yu2019open, hwang2018chatti} while there is a lack of tools that can help researchers and industrials focus on improving chatbots performance. We discuss this problem in section \ref{future}.
Our main contributions consist of :
\begin{itemize}
\item analyzing the current scientific and technological state of the art of conversational systems
\item presenting a new vision towards the standardization of conversational NLP 
\end{itemize}
\section{Chatbots categories}
We choose to categorize chatbots into two major types: social chatbots and task oriented chatbots.
\subsection{Social chatbots} They are designed to carry unstructured human-like conversations. They are considered as “chit-chat bots”. Currently, such systems may have an entertainment value but firstly, they were designed as a testbed for psychological therapy and they are still used, today, for this purpose. These systems (e.g., ELIZA \citep{weizenbaum_elizacomputer_1966}, PARRY \citep{colby_artificial_2013}, ALICE, CLEVER, Microsoft Little Bing, etc) have taken the first steps towards conversational agents.
\subsection{Task oriented chatbots} We also choose to classify them into two categories: generalist task oriented and specialist task oriented chatbots.
\subsubsection{Generalist task oriented chatbots}
A generalist chatbot is supposed to answer to general utterances. This feature is referred to as "chit-chat" and is designed to carry short conversations. Beyond this social aspect,  a generalist chatbot can also accomplish simple tasks related to everyday life such as setting the alarm, making a phone call, sending a text, etc. 
\subsubsection{Specialist task oriented chatbots} Specialist chatbots are designed for a very particular task. They are provided with a specific domain expertise that enables them to perform specific actions or to solve complex problems. A specific task can be to book a flight, to order food or even to analyze health problems.
\section{Chatbot building approaches}
We describe in this section two of the main chatbot building architectures: rule-based approches and data-driven approaches. rule-based systems were exploited. earlier and they rely on pattern-action rules. On the other side, we find the data-driven approaches which rely on large conversations corpora. 
\subsection{rule-based chatbots}
Rule-based chatbots are designed to answer questions based on prefixed rules. During the dialogue, the bot follows specific rules to chat with the user. For example, if the user utterance is part of [Hello", "Good morning", "Hi"] then the chatbot should answer with "Hello".  Rule-based chatbots are very popular in today's market as they are easy to build and performing for simple tasks. However, performing complex tasks requires writing many rules which can be time consuming. Despite their simplicity, the first proposed rule-based chatbots in history are impressing. ELIZA is a rule-based chatbot that simulates a Rogerian psychotherapist \citep{weizenbaum_elizacomputer_1966}. Its principle consists in applying pattern and transform rules. Each transform rule corresponds to a keyword. Keywords are ranked from specific to general with specific keywords being highly ranked. Then, in each user utterance, the chatbot finds the keywords with the highest rank in the knowledge base and applies the transform rule according to the sentence pattern.
Let's take the following sentence as an example of the user utterance.
\begin{center}
    \textit{'You hate me'}
\end{center}
This sentence matches the following pattern:
\begin{center}
(0 YOU 0 ME)
\end{center}
where 0 indicates a sequence of words with a variable length. We suppose that the keyword YOU is linked to the transform rule :
\begin{center}
    (0 YOU 0 ME) $\rightarrow$ (WHAT MAKES YOU THINK I 3 YOU?)
\end{center}
 where 3 refers to the third element in the pattern which is, in this case, the second zero and corresponds to the word \textbf{'hate'} in the source sentence. Applying this rule leads to the following answer:
 \begin{center}
     WHAT MAKES YOU THINK I \textbf{hate} YOU?
 \end{center}
 Following ELIZA, another system called PARRY specialized in psychotherapy has been proposed \citep{colby_artificial_2013}. PARRY uses rules similar to ELIZA but is more sophisticated as it has affect variables (e.g., anger, fear) that lead to more nuanced answers. These states are defined using formal logic rules.

\subsection{Data-driven chatbots}

Data-driven chatbots are the newest approaches and the most used ones. The availability of text datasets, in general, and conversational datasets, in particular, allowed to apply data-driven approaches to text. While rule-based approaches use hand-crafted rules to produce answers or even questions, data-driven approaches use existing human-human or bot-human conversations and/or narrative documents to create the bot utterances. In order to leverage the existing conversational data, one can use whether Information Retrieval (IR) or Machine Learning (ML). We will try, below, to report the most popular and recent data-driven approaches for chatbot building. 
\subsubsection{Information retrieval based chatbots}
Among the first IR chatbots, we can cite CleverBot. It was created by Rollo Carpenter in 1988 and published in 1997. CleverBot replies to questions by identifying how a human responded to the same question in a conversation database. Although it is simplistic, CleverBot was used a lot.
Information retrieval chatbot models work pretty much as a search engine where the query is the user turn and the search result is the chatbot answer. Having a Q-R\footnote{Question-Reply} pairs dataset and question Q, the IR based conversational model will look in the Q-R dataset for the pair (Q',R') that best matches Q and returns R’ as an answer to Q \citep{jafarpour_filter_2010,leuski_npceditor:_2011}. This can be considered as a way to mirror the training data.
To achieve these tasks, many retrieval baseline models have been proposed. Word-level Vector space models have been widely used with a cosine distance to find the best match Q-A pair \citep{banchs_iris:_2012}. Other works focused on using Term Frequency-Inverse Document Frequency (TF-IDF) retrieval models \citep{gandhe_surface_2013,charras_comparing_2016}. \citet{galitsky_chatbot_2017} proposed a chatbot for customer support and product recommendation that aims at helping users navigate to the exact expected answer as fast as possible. To this end, they suggest using discourse trees and particularly Rhetorical Structure Theory \citep{mann_rhetorical_1988} to model the generalization or specification relations between the possible answers.\\
Regarding the data sources for creating dialogue systems \citep{serban_survey_2015}, WikiAnswers\footnote{http://www.answers.com/Q/}, Yahoo Answers\footnote{https://answers.yahoo.com/?guccounter=1} and twitter conversations are among the most used open domain datasets for generalist IR based chatbots. Some researchers have gone beyond these datasets and applied the IR approach to narrative text datasets such as Wikipedia \citep{isbell_cobot_2000,yan_docchat:_2016}.
\subsubsection{Machine learning based chatbots}

\begin{figure}
\begin{center}
\includegraphics[scale=0.2]{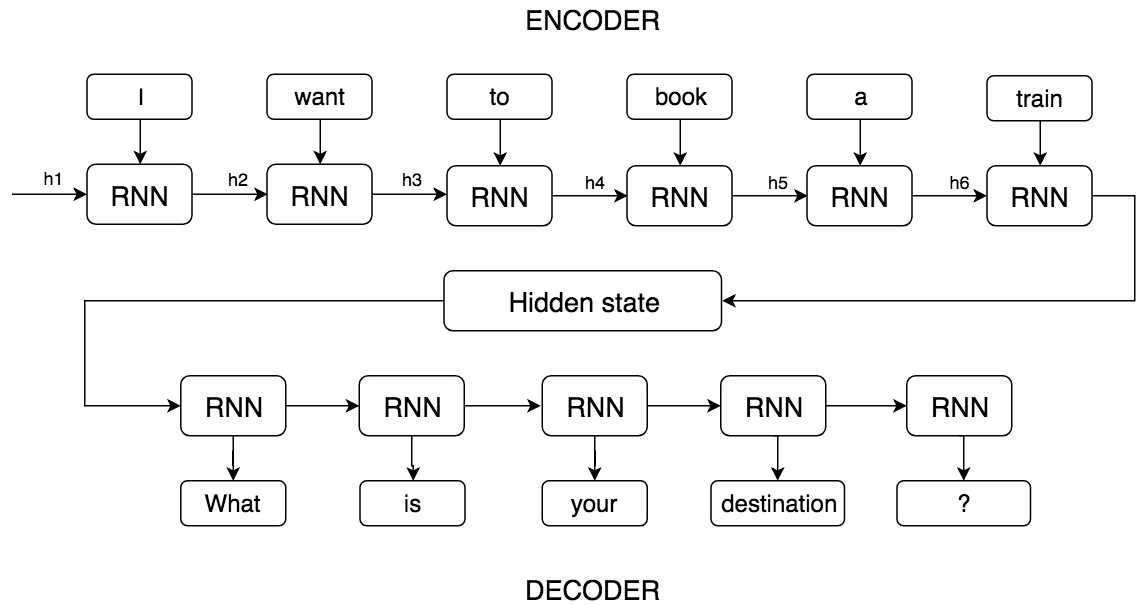}
\caption{Sequence to sequence architecture}
\label{img}
\end{center}
\end{figure}

The problem of generating human-like conversations has been modeled, lately, as the problem of mapping a human turn to a machine turn which is the target to be predicted. Most recent works focused on applying deep learning models but each of these formulates the problem in a particular way and uses a different set of features. Most used machine learning models are sequence-to-sequence learning and reinforcement learning. We describe them below.\\
\\
\textbf{Sequence to sequence learning}\\
Sequence to sequence (seq2seq) learning \citep{sutskever_sequence_2014} represents a pattern for using Recurrent Neural Networks (RNN) to tackle complex sequence-to-sequence prediction problems such as machine translation, image captioning \citep{park_attend_2017}, speech recognition \citep{chorowski_towards_2016}, text summarization \citep{rush_neural_2015,nallapati_abstractive_2016} and question-answering. Seq2seq learning have shown a great success when first applied on phrase to phrase machine translation \citep{cho_learning_2014} and therefore, inspired researchers to apply it for other tasks.\\
\indent For building chatbots, the problem was considered as translating the user utterance to the chatbot answer. Currently seq2seq models hold the state-of-the-art performance on chatbot building. These models are trained to map input sequences to output sequences. The length of the input and output sequences can be different and this is the strength of seq2seq models in comparison with other neural learning models. Technically, a seq2seq model is composed of an encoder and a decoder. The encoder is a neural network that  reads the input sequence and converts it into a hidden state called context vector or thought vector because it stores the meaning of the input sequence, considered as a thought. The decoder is then fed with this context vector. During the learning stage, it learns to map the hidden state to the true output sequence.  In the inference stage, the decoder returns the predicted output sequence with respect to the learning goal. Figure \ref{img} shows an example of a sequence to sequence architecture.\\
\citet{vinyals_neural_2015} proposed a straightforward data-driven approach where they trained a seq2seq model on large conversational datasets to predict the chatbot answer given the user question. Their work showed that despite the simplicity of the system, the seq2seq model was able to answer simple questions, extract relevant information from large corpus and perform shallow reasoning. However it is far from performing natural conversations. A similar model has been trained on TV shows and particularly on selected popular characters \citep{nguyen_neural_2017}. The trained chatbots inherited successfully the identity and the personality of the target character which shows that these models can capture many aspects. \citet{hu_touch_2018} proposed a modification to the seq2seq model so it captures the tone in customer care conversations (e.g., neutral, passionate, empathetic, etc.)  and takes it into account by generating toned answers. In the learning step, the seq2seq model is fed with not only pairs of user-agent turns but also with the corresponding tone of the first speaker. Considering the history as a context in seq2seq models led to some improvement over context-independent models but suffers from the highly general answers \citep{sordoni_neural_2015}.\\
\\
\textbf{Reinforcement learning}\\
  Reinforcement learning enables the machine to learn in the same way humans do, that is to say, by interacting with the surrounding environment. The learning process aims at maximizing a notion of cumulative reward. After each interaction, the agent observes the results of its action and receives accordingly a reward which can be positive or negative.\\
  \indent To train a dialogue system with reinforcement learning, the chatbot is put in use by the end users to become increasingly efficient throughout the conversations. Previously to the development of neural networks, the problem of dialogue systems have been modeled as Markov Decision Processes (MDPs) \citep{levin_using_1998} in order to use reinforcement learning. The system is represented by a set of states that correspond to the entire dialogue, and by a set of actions representing the system answers. The goal is to maximize the reward obtained for fulfilling user’s requests. This dialogue manager is preceded by an SLU (Speech Language Understanding) component and succeeded by an NLG (Natural Language Generation) component. \\
  \indent Later, dialogue systems have been modeled using Partially Observed Markov Decision Processes (POMDPs) \citep{roy_spoken_2000,young_talking_2002,williams_partially_2007,young_pomdp-based_2013,gasic_-line_2013}. This approach assumes that dialogue starts in an initial state $s_0$. Succeeding states are modeled by a transition probability $p(s_{t}\mid s_{t-1},a_{t-11})$ where $s_{t}$ is the state at time $t$ and $a_t$ is the action taken at time $t$ . The state $s_t$ is partially observable to take into account the error rate of the language understanding framework. To this end, at each turn, the user input is converted to an observation with probability $p(o_{t}\mid s_{t})$ where $o_t$ is the observation at time $t$. The transition and observation probability stochastic functions represent the dialogue model M.  The possible system actions are delegated to another stochastic model encoding the policy P. During the dialogue, a reward is assigned  to the system at each utterance in a way to favor the ideal behavior of the dialogue system. The dialogue model and the policy model are optimized by summing up the cumulated rewards during the dialogue. Reinforcement learning consists of learning the best policy through reward system. This process can be learned on-line or off-line. 

\subsubsection{Hybrid approaches}
Each of the described approaches has its own limits and strengths. In order to take the best part of each method, many studies focused on combining different approaches: machine learning and IR or seq2seq and reinforcement learning. 
Seq2seq generative models have been combined to IR approaches. This is the case of the Alibaba shopping assistant \citep{qiu_alime_2017} that uses an IR approach to retrieve the best candidates from the knowledge base then, the seq2seq model is used to re-rank the candidates and generate the answer. \citet{cui_superagent:_2017} propose a meta-engine customer service chatbot composed of four sub-engines. A fact QA engine answers questions about the product details found in the product page using a deep structured semantic model \citep{huang_learning_2013} for matching the question to the product attributes. A FAQ search engine maps the questions to the existing questions in the FAQ by  training a regression model using many semantic similarity measures. Another engine leverages the customers reviews to answer opinion-oriented questions. Finally, a chit-chat engine consisting of an attention based seq2seq model is trained on twitter conversations to answer greeting and thanking queries. \citet{li_deep_2016} highlighted the fact that seq2seq predict answers only one at a time and fail to predict their influence on the future utterances. They suggested a novel model that leverages the seq2seq ability to represent semantics and combines it to a reinforcement learning policy which optimizes long-term rewards according to the developer goal. Extending seq2seq models with history trackers and database information has also shown an improvement over basic seq2seq where only the user intent is decoded \citep{wen_network-based_2016}.

\section{Chatbots evaluation}

Chatbots can be evaluated through human evaluation. To this end, human judges are asked to score the chatbot performance on different criteria or to compare the chatbot answers to another chatbot answers. While it provides a high quality evaluation, human evaluation remains very expensive in time and human resources and cannot scale well.  Some IR based chatbots can be evaluated using simple precision and recall metrics as they operate as a search engine.
Recently , automatic evaluation approaches have been tested such as BLEU \cite{papineni2002bleu} and ROUGE \citep{lin_rouge:_2004} which are used respectively for automatic translation and text summarization. These metrics work by computing the n-gram overlap between the output of a system and a set of references. For chatbot evaluation, these metrics have been used to compare the chatbot answers to human answers to the same question. N-gram overlap based metrics has limits in dialogue systems evaluation as two answers can be totally different but have the same meaning.\\
Perplexity \cite{manning1999foundations} has also been used to evaluate chatbots \cite{vinyals2015neural,yao2016attentional,serban2017deep,cui2017superagent}. Used originally to evaluate language models, perplexity measures how well a language model predicts the probability of words in the test set. If we suppose that the test set is composed of $\left \{ {w_{1}, w_{2}, ... , , w_{N}}  \right \}$ the perplexity of the language model that should predict the probability of these words is:
\begin{equation}
    Perplexity = e^{-\frac{1}{N}\sum_{i}log(P_{w_{i}}))} 
\end{equation}
This means the lower the perplexity score the better.
More recent studies suggested new evaluation metrics inspired by the Turing test consisting of training a classifier to distinguish system answers from human answers \citep{li_adversarial_2017}.

\section{Discussion}

All the described methods have their own weaknesses and strengths. Choosing one approach or another depends deeply on the available data format for training (structured or unstructured). It depends also on the chatbot end-use. Some approaches may be performing for a chit-chat engine as a purely data-driven seq2seq model but not for a domain specific task oriented chatbot. rule-based approaches are straightforward  but efficient for simple use cases. They can also be used jointly with data-driven approaches to handle mandatory scenarios that aren’t present in the knowledge base. Rules can also be used to support data-driven methods to guarantee a security level, especially, in industrial cases where machine learning, on its own, cannot be fully trusted. IR methods are the best when it comes to selecting the best piece of information from structured data to respond to a user query. While IR methods can be very performing in many cases, they need an extra component that adapts the form of the retrieved/selected information to be suitable as an answer. Moreover, purely IR based methods don't perform reasoning and are then perfect to mirror existing knowledge, not more. \\\\
\indent Machine learning based approaches are the dominating methods now in chatbot building. While reinforcement learning has been the main used learning method in dialogue systems for years, the encoder-decoder learning is now taking the lead. Reinforcement learning has been widely used in robotics as it allows the robot to be autonomous. The same thing is noticed in chatbot systems. Reinforcement learning leads to a more natural chatbot as it learns from human feedback and develops its own control system.  Reinforcement learning enables the chatbot to handle a long conversation and takes into account the preceding turns. It is also convenient to use RL to make chatbots learn to act and not only to chat. For example, a flight booking agent will receive a reward if it properly books a flight according to the user request and will get penalized if it makes an error. However, RL raises two main issues. Firstly, it needs so much time and so many interactions until the agent is trained which may be restricting if the training is online. Secondly, RL is not most suitable option to learn language generation. That is the reason behind the glory of the generative seq2seq models that have partially solved the problem. Seq2seq models are, usually, trainable end-to-end without any hand-crafted rules. Despite their success, seq2seq models have also some weaknesses. The basic seq2seq models are effective however they are more performing in generalist chatbots than in task oriented chatbots. Another problem is that these models tend to return very generic answers \citep{sordoni_neural_2015} because of the high number of generic sentences in the  training data and the objective function of the seq2seq models. Besides, these models don't take repetition into account. For example, if the chatbot says "goodbye"
and the user replies with « goodbye", than it is possible that the chatbot answers again with a "goodbye". This is a simple example that can be handled easily but it is just to show how the chatbot can be stuck in a loop. Hybrid systems that combine RL with seq2seq or those who extend seq2seq models with external information (e.g., a database) seem to overcome some of the listed problems or at least to produce a more performing chatbot. 

\section{Opinion piece : towards the standardization of the field}\label{future}

Recently, AI researchers have started to pay more attention to conversational agents and attempts to build strong chatbot systems are being published increasingly. Since this field is not technologically mature we can obviously notice the lack of standard open source tools or models that can help scientists save time on the implementation step. For example, there is a common part among all the chatbot systems : the chatbot engine. This part represent the backbone of the chatbot system and should be independent from the used approach (ML or rule-based). Unfortunately, the distinction between the generic chatbot engine and the conversational NLP protocol is not always established. As a consequence,  a repetitive work is done by each member of the community by building chatbot systems from scratch every time an approach is proposed. While some code blocks are available to reuse, one should invest time to understand the code and adapt it to his CNLP techniques. This process slows down the progress in the chatbot field. If we take a look at other research fields we can observe the establishment of some well known standards that allowed to make huge steps on the related field. For example, the available open source standards for Machine Learning and Deep Learning techniques (e.g., TensorFlow \cite{abadi2016tensorflow}) allowed the researchers to explore a multitude of applications and that has contributed to many advances in different fields. By analogy, we can say that there's a need to such an open source standard for the CNLP field. Considering these facts, we think future studies on chatbot systems should focus on setting up standard frameworks and architectures for conversational agents. Such standards should be interoperable and should allow NLP researchers to plug their NLP in it.

\section{Conclusion}

In the light of these observations, we believe that developing a natural conversational agent cannot be solved with one simple model. As was the case with many problems, system aggregation usually leads to better results than each system alone. Future works should maybe focus on finding the best way to combine the different approaches to chatbot building in order to take better advantage of each method, even the simplest ones.\\
On the other hand, the diversity of the approaches for chatbot building and the growing interest to this field suggest to start thinking about setting standard tools and platforms to boost the development of this domain and help the scientific community focus on new models.
\bibliographystyle{acl_natbib_nourl}
\bibliography{emnlp2018}

\end{document}